\documentclass[preprint,authoryear]{elsarticle}

\usepackage{lineno,hyperref}
\usepackage{tabularx}
\usepackage{array}
\usepackage{amsmath}
\usepackage{multirow}
\usepackage{mathrsfs}
\usepackage[usenames, dvipsnames]{xcolor}
\usepackage{graphicx}
\usepackage{stmaryrd}
\usepackage{amssymb}
\usepackage{bigstrut}
\usepackage{color,colortbl}
\usepackage{tabu}
\usepackage{pdflscape}
 \usepackage{array,multirow,graphicx}
 \usepackage{float}

\usepackage{breakurl}

\newcommand*\concat{\mathbin{\|}}
\newcolumntype{L}[1]{>{\raggedright\let\newline\\\arraybackslash\hspace{0pt}}m{#1}}
\newcolumntype{C}[1]{>{\centering\let\newline\\\arraybackslash\hspace{0pt}}m{#1}}
\newcolumntype{R}[1]{>{\raggedleft\let\newline\\\arraybackslash\hspace{0pt}}m{#1}}
\modulolinenumbers[5]
\journal{}
\emergencystretch=\maxdimen
\begin{document}

\definecolor{rev1color}{rgb}{0, 0, 0}	
\definecolor{rev2color}{rgb}{0, 0, 0}

\begin{frontmatter}
\title{Density-Wise Two Stage Mammogram Classification using Texture Exploiting Descriptors\tnoteref{mytitlenote}}
\tnotetext[mytitlenote]{This material is based upon work supported by Council of Scientific and Industrial Research (India) Grant Number 25/(0220)/13/EMR-II.}

\author[mymainaddress]{Aditya A. Shastri}
\ead{phd1501201001@iiti.ac.in}

\author[mymainaddress]{Deepti Tamrakar}
\ead{deeptitm@iiti.ac.in}

\author[mymainaddress]{Kapil Ahuja\corref{mycorrespondingauthor}}
\cortext[mycorrespondingauthor]{Corresponding author}
\ead{kahuja@iiti.ac.in}
\ead[url]{http://iiti.ac.in/people/~kahuja/}
\address[mymainaddress]{Computer Science and Engineering, Indian Institute of Technology Indore, Simrol, Indore, MP 453552, India}

\begin{abstract}
Breast cancer is becoming pervasive with each passing day. Hence, its early detection is a big step in saving the life of any patient. Mammography is a common tool in breast cancer diagnosis. The most important step here is classification of mammogram patches as normal-abnormal and benign-malignant. 

Texture of a breast in a mammogram patch plays a significant role in these classifications. {\color{rev2color}We propose  a variation of Histogram of Gradients (HOG) and Gabor filter combination called Histogram of Oriented Texture (HOT) that exploits this fact.} We also revisit the Pass Band - Discrete Cosine Transform (PB-DCT) descriptor that captures texture information well. All features of a mammogram patch may not be useful. Hence, we apply a feature selection technique called Discrimination Potentiality (DP). Our resulting descriptors, DP-HOT and DP-PB-DCT, are compared with the standard descriptors. 

Density of a mammogram patch is important for classification, and has not been studied exhaustively. The Image Retrieval in Medical Application (IRMA) database from RWTH Aachen, Germany is a standard database that provides mammogram patches, and most researchers have tested their frameworks only on a subset of patches from this database. We apply our {\it two new} descriptors on {\it all} images of the IRMA database for {\it density wise classification}, and compare with the standard descriptors. We achieve higher accuracy than all of the existing standard descriptors (more than 92\%).

 %
\end{abstract}
\end{frontmatter}

\section{Introduction}\label{secIntro}
Breast cancer has become the most common killer disease in the female population. Collectively India, China and US have almost one-third burden of global breast cancer \citep{static}. The abnormalities like the existence of a breast mass, change in shape, the dimension of the breast, differences in the color of the breast skin, breast aches, etc., are the symptoms of breast cancer. Cancer diagnosis is performed based upon non-molecular criteria like the tissue type, pathological properties and the clinical location. Cancer begins with the uncontrolled division of one cell and results in the form of a tumor.

There are several imaging techniques for examination of the breast, such as magnetic resonance imaging, ultrasound imaging, X-ray imaging, etc. Mammography is the most effective tool for early detection of breast cancer that uses a low-dose X-ray radiation. 
It can reveal pronounce evidence of abnormalities, such as masses and calcification, as well as subtle signs, such as bilateral asymmetry and architectural distortion. The diagnosis of breast cancer by classifying it as benign and malignant in the early stage can reduce chances of the death of the patient. 

Mammographic Computer Aided Diagnosis (CAD) systems enable evaluation of abnormalities (e.g., micro-calcification, masses, and distortions) in mammography images. CAD systems are necessary to aid facilities in carrying out a more accurate diagnosis. CAD systems are designed with either fully automatic or semi-automatic tools to assist radiologists for detection and classification of mammography abnormalities \citep{oliver2010review}. In semi-automated CAD systems, enhancement techniques are first applied on a mammogram patch, radiologists then select a Region of Interest (ROI) or a patch, and finally, the patch is classified by the system.

Mammogram patch classification is often done in one stage. However, classifying a mammogram patch in multiple stages is also beneficial. Two-stage classification of mammogram patches helps in reducing the possibility of a false positive classification. In the first stage, mammogram patches are classified as normal or abnormal (mass), then in the second stage, abnormal patches are further classified into benign or malignant. This work proposes two-stage mammogram patch classification. The system is trained with normal, benign and malignant mammogram patches separately.

Generally, CAD systems consist of basic modules as follows: mammogram patch pre-processing, breast segmentation, enhancement, feature extraction and classification \citep{rangayyan2007review}. Pre-processing step helps in removal of irrelevant regions present in a mammogram patch such as pectoral muscles and digit information.
Breast region is segmented using a threshold. Enhancement techniques such as adaptive histogram equalization, non-linear filtering are applied on the breast region to improve visualization of tissues or a tumor in a mammogram patch \citep{anand2015mammogram, sundaram2011histogram, Jenifer2016contrast}. In most works, shape features of a mammogram patch have only been considered. The shape of a mammogram patch plays an important role for benign and malignant classification. While benign masses have round or oval shapes with clear margins, malignant masses with spicule have jagged edges \citep{mudigonda2000gradient}. Appropriate features of mammogram patches help in accurate classification.

Mammogram patches can be better classified by using their texture properties\footnote{It is relatively hard to classify a mammogram patch as benign-malignant (as compared to normal-abnormal) due to their lack of distinctive properties \citep{mudigonda2001detection}.}. {\color{rev2color}This work proposes a descriptor that captures the textural features of a mammogram patch, i.e. Histogram of Oriented Texture (HOT), which is a variation of Histogram of Gradients (HOG) and Gabor filter combination.} We also apply the existing Pass Band - Discrete Cosine Transform based descriptor (PB-DCT) here because of its advantage in helping filter textural features. These descriptors have not been used yet for mammogram patch classification. We use Discrimination Potentiality (DP) to select appropriate features of mammogram patches in these two descriptors, resulting in two new descriptors. The proposed descriptors are compared with the {\color{rev1color}six standard descriptors} for mammogram patch classification; Zernike moments \citep{tahmasbi2011classification}, {\color{rev1color}MLPQ \citep{nanni2012very}, GRsca \citep{nanni2013different},} Wavelet Gray Level Co-occurrence Matrix (WGLCM) \citep{beura2015mammogram}, Local Configure Pattern (LCP) and Histogram of Gradients (HOG) \citep{ergin2014new}. SVM is the most suitable classifier for two-class classification and is widely used in this field. Hence, we use this.    

Breasts with high density have a higher chance of cancer. However, high dense tissues and masses appear as mostly white in a gray scale of a mammogram patch. Hence, it is very difficult to detect a tumor in high dense tissues. Especially, the difference between benign and malignant tumors is hard to determine \citep{Oliveira2011,oliver2008novel,oliver2012automatic,oliver2010review,petroudi2011breast}. 
Generally, breasts are classified based upon density in three different ways by the Breast Imaging Reporting And Database Systems (BIRADS); two classes (fatty and dense), three classes (fatty, glandular, and dense) or four classes (mostly fatty, scattered density, consistent density and extremely dense) \citep{Oliveira2011,oliver2008novel}. 
Most researchers in this area have not considered the density of a breast for mammogram patch classification (normal-abnormal and benign-malignant). Hence, in this work, 
using a two-stage mammogram patch classification system, we test our two proposed descriptors for each BIRADS class separately. 

CAD systems are usually tested on the MIAS and DDSM mammogram patch datasets of the IRMA database \citep{IRMA}. The MIAS dataset consists of a small set of images, while DDSM includes few thousand images. Several descriptors and methodologies have been proposed for mammogram patch classification, but their performances have been investigated only for a small set of images. Moreover, these systems have not achieved desired accuracy \citep{petroudi2011breast,Oliveira2011}. The performance of our system is tested on all mammogram patches of the MIAS and DDSM datasets. The experimental results show the effectiveness of our approach; we achieve near to 92\% accuracy.


The rest of this paper is organized as follows. Section \ref{sec:lit}  provides a summary of the related work. Section \ref{sec:propose} explains the proposed CAD system. Section \ref{sec:exp} presents experimental results. Finally, Section \ref{sec:summary} gives conclusion and discusses future work.

		 
\section{Literature Review}
\label{sec:lit}
Researchers have reviewed existing techniques for detection and analysis of abnormalities in mammogram patches as discussed earlier (calcification, masses, tumors, bilateral asymmetry, and architectural distortion) \citep{rangayyan2007review}. Some people have also reviewed the contribution of texture to risk assessment for each density separately \citep{oliver2010review}. Mammogram patches consist of directionally oriented, texture image due to its fibro-glandular tissues, ligaments, blood vessels and ducts. 
These texture features for mammogram patches can be categorized into four groups; statistical \citep{oliver2012automatic, rabottino2008mass, shanthi2012computer, peng2016automated, beura2015mammogram}, local pattern histogram \citep{abdel2015analysis,ergin2014new,Oliveira2011,petroudi2011breast}, directional \citep{buciu2011directional,jasmine2009microcalcification, gedik2016new, mudigonda2000gradient,mudigonda2001detection}, and transform based \citep{Laadjel2015Biometric,Dabbaghchian2010Feature}.

Statistical features such as mean, variance, energy, entropy, skewness, and kurtosis are mostly utilized as a descriptor for classification \citep{oliver2012automatic, rabottino2008mass, shanthi2012computer, peng2016automated, beura2015mammogram}. Gray Level Co-occurrence Matrix (GLCM) and Gray Level Run Length Matrix (GLRLM) provide the relationship between neighboring pixels of a mammogram patch. Statistical properties of these matrices have also been exploited for mammogram patch classification \citep{beura2015mammogram}. These features are extracted by directly using spatial data from images.

Some works have exploited local distribution of textural properties of mammogram patches for classification. Histogram of Gradients (HOG) \citep{ergin2014new}, 
Local Configure Pattern (LCP) \citep{ergin2014new}, Uniform Directional Pattern (UDP) \citep{abdel2015analysis}, Local Ternary Pattern (LTP) \citep{Muramatsu2016Breast}, {\color{rev2color}Local Phase Quantization (LPQ) \citep{ojansivu2008blur} and Local Binary Pattern (LBP) \citep{oliver2007false} are some such examples. There are three different variants of LBP, which are usually used for exploiting local textural properties; Uniform Local Binary Pattern (LBP-u), Rotation Invariant Local Binary Pattern (LBP-ri) and Rotation Invariant Uniform Local Binary Pattern (LBP-riu) \citep{nanni2012very, ojala2002multiresolution}.} Block-wise feature extraction gives better performance as compared to global feature vectors. Statistical properties of local histogram have also been used as a mammogram patch descriptor for classification \citep{wajid2015local}.  

Coming to directional features, wavelet, dual-tree complex wavelet Gabor, Contourlet, finite Shearlet, etc. have been exploited for multi-resolution and multi-orientation texture or tissue analysis of a mammogram patch 
\citep{jasmine2009microcalcification, gedik2016new, mudigonda2000gradient,mudigonda2001detection}. Gabor based feature extraction schemes are widely used for mass classification as benign-malignant. Gabor features can be extracted from mammogram patches in different ways \citep{buciu2011directional,khan2016optimized}. Recently, some people have proposed directional features of mammogram patches computed by a Gabor wavelet for four different scales and eight different orientations \citep{buciu2011directional}. Optimal parameters of a Gabor filter increases discrimination between normal and abnormal properties.

Finally, the fourth category for texture feature extraction is by using a mathematical transform. For example, Discrete Cosine Transform is one such option \citep{Laadjel2015Biometric,Dabbaghchian2010Feature}.

In this work, we first propose a descriptor that exploits local distribution of textural property (HOG) as well as considers directional features (Gabor). We term it as Histogram of Oriented Texture (HOT). \textit{The reason for deriving this descriptor is that, for density-based mammogram patch classification, individually these two descriptors have their own drawbacks, which are eliminated in their combination.} The width of tissues may vary with the density of a mammogram patch, and it is difficult to estimate with HOG.  Applying a Gabor filter for feature extraction on the whole mammogram patch is not useful since abnormalities are usually very local. {\color{rev2color} The combination of HOG and Gabor filter is not new (see \cite{conde2013hogg, ouanan2015gabor, xu2015facial}). However, none of these works have applied this combination to mammogram patch classification. Moreover, our combination is optimally designed for solving the problem at-hand. We do a detailed comparison of our descriptor with these existing ones at the end of Section \ref{sec:hot}, i.e. after describing our descriptor.}   

 
The HOT descriptor mentioned above has few drawbacks in terms of capturing all textural features. Next, we revisit a transform based descriptor; Pass Band - Discrete Cosine Transform (PB-DCT). This descriptor has not been used yet for density-based mammogram patch classification. DCT has very strong energy compaction capability, i.e., an image can be represented by a small set of coefficients. DCT coefficients are divided into three bands; high, middle and low. The high-frequency coefficients correspond to irrelevant information, the medium frequency coefficients carry textural information, and the low-frequency coefficients contain illumination information. We use PB-DCT like a band pass filter to extract mostly the middle frequencies with some amount of low frequencies as well.

The dimension of features can be reduced by either using a feature selection scheme or a dimension reduction scheme \citep{gedik2016new}. Feature selection schemes select the appropriate feature set based upon a criterion (such as entropy, fisher, maximum mutual information, etc.), while dimension reduction schemes project features onto an other dimensional subspace using orthogonal matrices. If the number of training samples for each class is less than the dimension of features, it is known as small sample size (SSS). Under this circumstance, which is common, some matrices in the dimension reduction approach become singular leading to difficulty in further computation \citep{gedik2016new}. In general, it has been found from literature that the rank-based feature selection approach is more suitable for feature reduction. Hence, we use this. Genetic algorithm can also be utilized for selecting suitable features from intensity, texture and shape features for benign and malignant classification of mammogram patches \citep{rouhi2015benign}. This is part of future work.    

Support Vector Machine (SVM), K-Nearest Neighbor (KNN), Fisher Linear Discriminant (FLD), Naive Bayes and Neural Network with Multi-Layer Perception Learning have been utilized for mammogram patch classification \citep{gedik2016new, beura2015mammogram, wajid2015local}. SVM is the most suitable classifier for two class classification and is widely used in this field. Hence, we use this. {\color{rev2color}Recently, ensembles of two or more classifiers (also called as multiclassifiers) have been used to improve the classification accuracy (see \cite{nanni2012very}). In this work, first, different descriptors are obtained by varying certain parameters. Then, after extracting features by using each of these descriptors, different classifiers are trained. Finally, these classifiers are combined by some technique (e.g., different SVMs are combined by a sum rule).} This type of work will be explored in future.

Mammogram patches are usually categorized into four classes based upon the level of density (i.e., fat transparent ($d$), fibro-glandular ($e$), heterogeneously dense ($f$), and extremely dense ($g$)). As earlier, this is called the BIRADS classification. Each BIRADS category is divided into three classes as normal, benign and malignant \citep{de2008breast,IRMA,deserno1texture}.

\begin{figure}[h!]
	\centering
		\includegraphics[width=.8\textwidth]{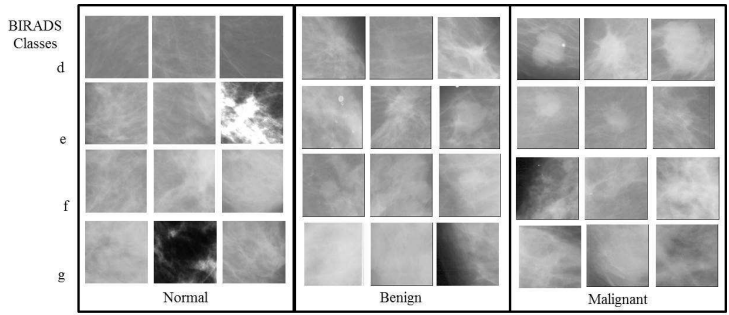}
	\caption{	\label{fig:Patches} Samples of mammogram patches from the IRMA database. Row denotes the density of patches. } 
\end{figure}

The IRMA reference database is a repository of mammogram patches and has been created by Deserno et al. to test the accuracy of approaches for mammogram patch classification. It contains the two datasets, MIAS and DDSM. This database provides information about images based on the type of background tissue and the class of abnormality present in the mammogram patch. Table \ref{tab:Irma_classes}  lists the number of images in both the MIAS and DDSM datasets based upon the above discussed classification. Some sample patches are shown in Fig. \ref{fig:Patches}

\begin{table}[h!]
\centering
\caption{\label{tab:Irma_classes} Distribution of normal, benign, and malignant mammogram patches of the two different datasets for the four BIRADS classes.}\smallskip
\begin{tabular}{|c|c|c|c|c|}
\hline
\multicolumn{5}{|c|}{IRMA: MIAS Patch Dataset}\\\hline
BIRADS&  	Normal&  	Benign& 	Malignant& 	Total\\\hline 
$d$& 	12& 	14& 	11& \textbf{37} \\\hline
$e$& 	28& 	1& 	5& \textbf{34}\\\hline
$f$& 	24& 	8& 	6& 	\textbf{38}\\\hline
$g$& 	26& 	9& 	6& 	\textbf{41}\\\hline
Total& 	\textbf{90}&	\textbf{32}&	\textbf{28}&	\textbf{150}\\\hline
\multicolumn{5}{|c|}{IRMA: DDSM Patch Dataset}\\\hline
$d$& 	203& 	219& 	222& 	\textbf{644}\\\hline
$e$& 	168& 	232& 	228& \textbf{628}\\\hline
$f$& 	195& 	225& 	227& 	\textbf{647}\\\hline
$g$& 	207& 	224& 	226& 	\textbf{657}\\\hline
Total& 	\textbf{773}&	\textbf{900}&	\textbf{903}&	\textbf{2576}\\\hline
\end{tabular}
\end{table}

Some of the related works are summarized in Table \ref{tab:lit}, along with the details of feature vectors, classifiers, and a number of images used in experiments. The accuracy obtained for both types of classification (normal-abnormal and benign-malignant) is also given. The performance of approaches depends on different factors of mammogram patches such as dimension, number of training and testing samples, resolution and the type of abnormality. Most of the works in this area have tested on a subset of images instead of all mammogram patches, which we use.

\begin{table*}[h!]
\caption{\label{tab:lit}Summary of some related works on mammogram patch classification.}\smallskip
\begin{tabular}{@{}|L{3.7cm}|p{1.3cm}|p{1.3cm}|p{1.3cm}|R{1.5cm}|R{1.5cm}|@{}}\hline
 \multirow{2}{*}{}{Feature} &\multirow{2}{*}{}{Classifier} & \multirow{2}{*}{}{Database} & Number  & Accuracy  & Accuracy  \\ 
  &  &   & of images &  (normal-abnormal)&  (benign-malignant)\\\hline
Gabor + PCA \citep{buciu2011directional}&SVM&DDSM &NA&84.00\%&78.00\%\\\hline

 GLCM + DWT &\multirow{2}{*}{}{BPNN$^*$}&MIAS&332&98.10\%&95.04\%\\\cline{3-6}
 \citep{beura2015mammogram} &  &	DDSM&550 &99.45\%&97.61\%\\\hline

HOG, DSIFT, \& LCP \citep{ergin2014new} &SVM&DDSM &600&84.00\%&78.00\%\\\hline

FFST &\multirow{2}{*}{SVM}&MIAS&228 &98.29\%&100.00\%\\\cline{3-6}
\citep{gedik2016new}&&DDSM&228  &100.00\%&98.29\%\\\hline

Gabor & PSO$^{**}$+ &\multirow{2}{*}{}{DDSM} & \multirow{2}{*}{}{1024}& \multirow{2}{*}{}{98.82\%}&\multirow{2}{*}{}{91.61\%}\\
 \citep{khan2016optimized} &  SVM&  &  &  & \\\hline 
\end{tabular}
\smallskip \smallskip \\ \footnotesize{$^*$BPNN: Back Propagation Neural Network. \\ $^{**}$PSO: Particle Swarm Optimization.}
\end{table*}

To summarize, in this work the two proposed feature extraction descriptors (HOT and PB-DCT with feature selection technique called Discrimination Potentiality) are compared with existing techniques for two-stage classification of all images of the IRMA database. This is done for individual density as well as together. We achieve higher accuracy than existing techniques for all images.

\section{Proposed Mammogram Patch Classification System}
\label{sec:propose}
\begin{figure}[h!]
\centering
\begin{tabular}{cc}
\includegraphics[width=0.45\textwidth]{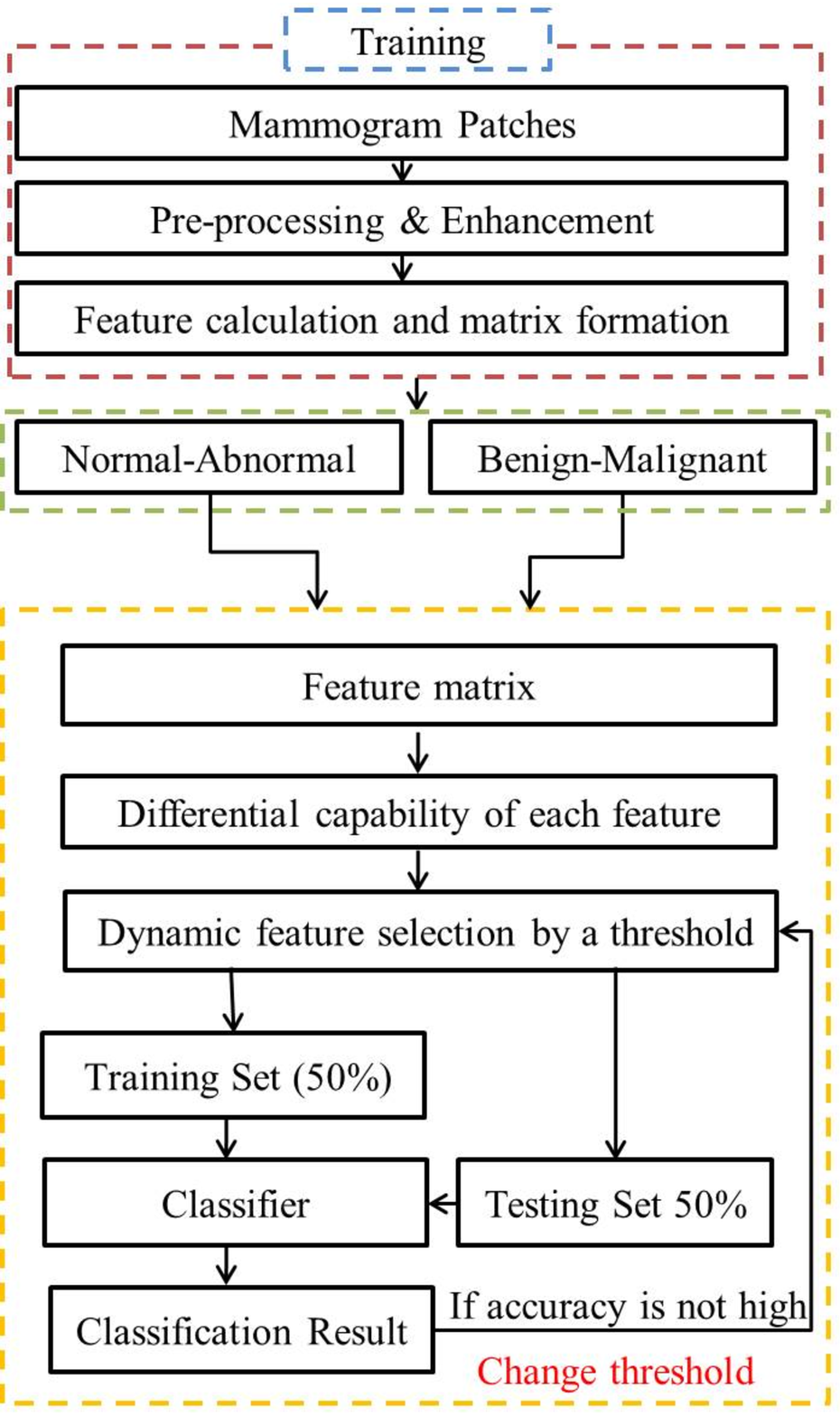}&\includegraphics[width=0.35\textwidth]{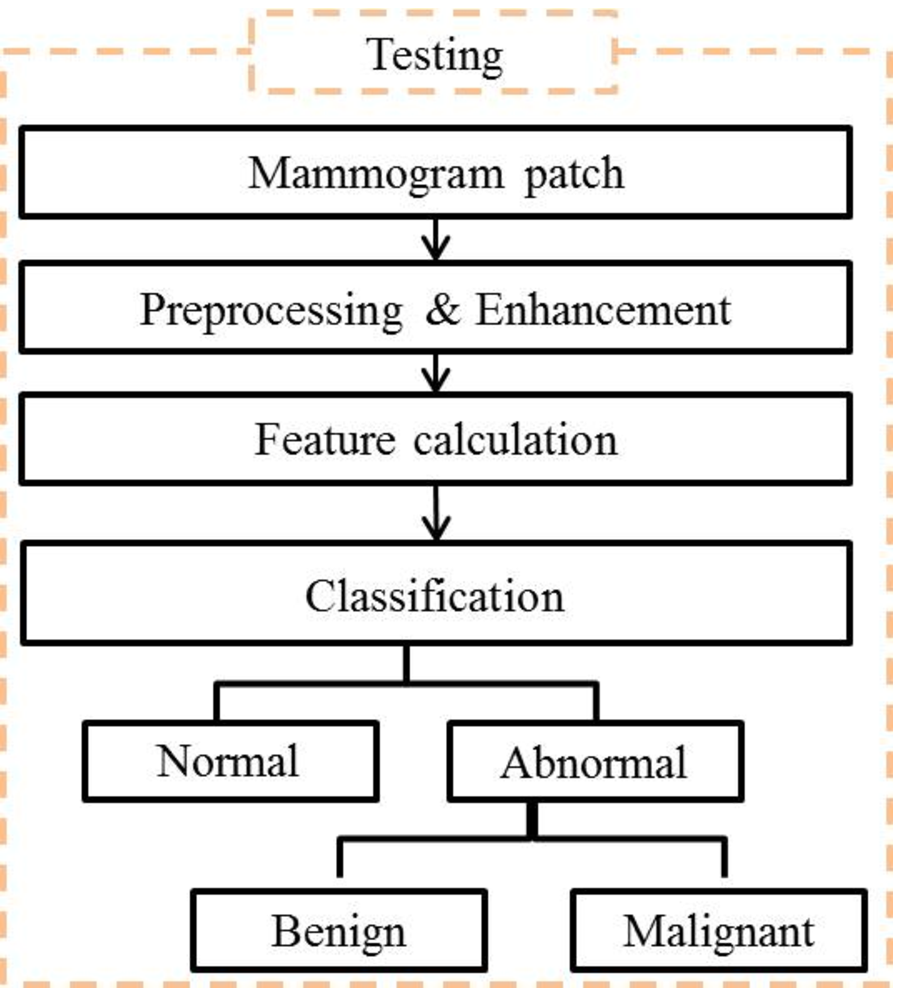}
\\ (a) & (b)
\end{tabular}
\caption{\label{fig:flowdiagram} Flow diagram of the proposed mammogram patch classification system; (a) training phase and (b) testing phase.}
\end{figure}

This work proposes a two-stage mammogram patch classification system. In the first stage, mammogram patches are classified as normal-abnormal, and in the second stage, abnormal mammogram patches are classified as benign-malignant. The framework of the proposed work for training and testing phase is shown in Fig. \ref{fig:flowdiagram}.

Here, we first discuss image pre-processing and enhancement techniques used by us. Mammogram patches are preprocessed for illumination normalization and visibility enhancement of tumors and tissues \citep{anand2015mammogram,sundaram2011histogram}. A two-stage adaptive histogram equalization enhancement technique is used here for texture enhancement of mammogram patches \citep{anand2015mammogram}. Second, we discuss our two proposed feature extraction techniques, where features of mammogram patches are extracted from enhanced images. Finally, we discuss the feature selection technique used by us. 


\subsection{Pre-processing and Enhancement}\label{subSecPreprocess}
\begin{figure}[h!]
	\centering
		\includegraphics[width=0.50\textwidth]{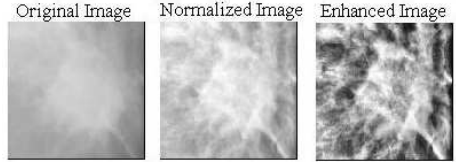}
\caption{\label{fig:preprocessed} A preprocessed and enhanced mammogram patch.}
\end{figure}

{\color{rev2color}In this work, for pre-processing, we only normalize the intensity of pixels and that too for a few mammogram patches. This is because while capturing images, illumination conditions are usually not the same. So, the range of the gray level is different for different mammogram patches. Hence, we use a simple and the most commonly used normalization formula (given below), which normalizes the intensity of pixels between 0 and 1 \citep{jain2005score, srisuk2008gabor}.
\begin{equation}
I'(x,y) = \frac{I(x,y)-\min(I)}{\max(I)-\min(I)}, \nonumber
\end{equation}
where $(x, y)$ is the pixel position, $I'(x, y)$ is the normalized pixel intensity, $I(x, y)$ is the actual pixel intensity, $\min(I)$ is the minimum intensity over all the pixels, and $\max(I)$ is the maximum intensity over all the pixels.

Next, we discuss tissue enhancement of mammogram patches. Histogram equalization is the one of the most basic technique here, which stretches the contrast of the high histogram regions and compresses the contrast of the low histogram regions. As a result, if the region of interest in an image occupies only a small portion, it will not be properly enhanced during histogram equalization. 

This leads to more advanced techniques for enhancement, e.g., Adaptive Histogram Equalization (AHE), Contrast Limited Adaptive Histogram Equalization (CLAHE), Unsharp Masking (UM), Non-Linear Unsharp Masking (NLUM), Two-Stage Adaptive Histogram Equalization (TSAHE), etc \citep{anand2015mammogram, panetta2011nonlinear}. 

CLAHE has been found to be more suitable for tissue enhancement in mammogram patches. One aspect of enhancement is to capture the texture of the breast in mammogram patches better, which is defined by entropy. In the work of \cite{anand2015mammogram}, authors have shown that TSAHE performs better than most of the existing techniques in not just the overall enhancement (defined by EME, i.e. Measure of Enhancement) but entropy as well. 

Since cancerous cells mostly develop in tissues and our proposed descriptors (HOT and PB-DCT) are strongly tied to breast texture, we use a combination of CLAHE and TSAHE. We term it as TS-CLAHE. 

We apply two stages of CLAHE on mammogram patches in a cascaded order.} Firstly, histogram equalization is applied to $8 \times 8$ sized blocks, followed by an application to $4 \times 4$ sized of blocks. Fig. \ref{fig:preprocessed} shows the normalized and the enhanced image of a mammogram patch. It is observed that mass tissues are clearly visible in the enhanced image.

\subsection{Feature Extraction Techniques}\label{subSecExtraction}

As discussed in Sections \ref{secIntro} and \ref{sec:lit}, this work proposes two descriptors (HOT and PB-DCT) for mammogram patch classification. 
HOT is a modification of the HOG descriptor where a Gabor filter is used to calculate the angle and the magnitude response of texture of a mammogram patch. Selected PB-DCT coefficients based features are used here to improve the classification accuracy for each density class. Next, we discuss these two techniques separately. To the best of our knowledge, these strategies have not been applied anywhere for mammogram patch classification.

\subsubsection{The Histogram of Oriented Texture (HOT) Descriptor}
\label{sec:hot}
 Here, we derive our HOT descriptor. Firstly, we discuss the calculations of gradient and orientation of an image as well as the HOG descriptor calculation from cells and blocks partitions \citep{ergin2014new}. Secondly,  we describe a Gabor filter, which is used to extract magnitude and orientation of tissue texture information, and finally, we discuss modifications to the HOG descriptor that involves a Gabor filter and parameter selection.

Gradient of an image $I$ in horizontal and vertical directions, for a pixel position $(x,y)$ is computed as   
\begin{eqnarray}
dx& = &I(x+1,y)-I(x-1,y) \textrm{ and} \nonumber\\
dy&=&I(x,y+1)-I(x,y-1), \nonumber
\end{eqnarray}
respectively. For each pixel, $I(x,y)$, the gradient magnitude $m(x,y)$ and orientation $\theta(x,y)$ are computed as below.
\begin{eqnarray}
m(x,y) &=& \sqrt{dx^2+dy^2} \textrm{ and} 
\label{eq:m_hog} \\
\theta(x,y)&=&tan^{-1}\left(\frac{dy}{dx}\right).
\label{eq:t_hog}
\end{eqnarray}

\begin{figure}[h!]
	\centering
		\includegraphics[width=.4\textwidth]{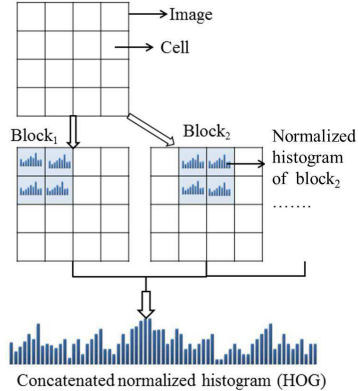}
	\caption{The HOG descriptor calculation.}
	\label{fig:cell}
\end{figure}

Orientation range ($0^{0}-180^0$) is quantized into $B$ bins (i.e.,  $\theta(x,y) \in bin(b)$ with $b = 1,2,3,\ldots,B$). The image is divided into $c\times c$ non-overlapping cells, and $l \times l$ cells are integrated as one block. Two adjacent blocks can overlap. The histogram of orientations ($HC(b)_{i}$) of bin$(b)$ within $i^{\textrm{th}}$ cell is computed as
\begin{eqnarray}
HC(b)_{i}&=&HC(b)_i+m(x,y),\nonumber \\
m(x,y)& \in& Cell_{i}, \nonumber  \\
b&=& 1,2,3,\ldots,B,\ \text{and}\nonumber \\
i&=& 1,2,3,\ldots,c\times c. \nonumber
\end{eqnarray}
The histogram of $j^{\textrm{th}}$ block ($HB_{j}$) is obtained by integrating HCs (Histogram of Cells) within this block as follows:
\begin{eqnarray}
{HB}_{j} &=& HC_{1}\concat
HC_{2}\concat
\ldots\concat
HC_{l \times l },
\nonumber 
\end{eqnarray}
where $\concat$ denotes histograms concatenation into a vector.  The vector of $HB_{j}$ is finally normalized by $L_2$-norm block normalization as below to obtain ${NHB}_{j}$.
\begin{equation}
{NHB}_{j} = \frac{HB_{j}}{\sqrt{\|HB_{j}\|^2_2+e^2}},\nonumber
\end{equation}
where $e$ is a small constant to avoid problem of division by zero. Histogram of Oriented Gradients (HOG) can be obtained by integrating normalized histograms of all blocks as below.

\begin{equation}
HOG ={NHB_{1} \concat NHB_{2} \concat \ldots NHB_{j}\concat\ldots,\concat NHB_{N}},  \nonumber
\end{equation}
 where $N$ is the number of possible blocks in an image, which is equal to $(c-l+1)\times (c-l+1)$.  
Fig. \ref{fig:cell} shows an example of cell partitions, formation of overlapped blocks and concatenation of histograms to get the HOG descriptor. Finally, the length of HOG is $l^2 \times {(c-l+1)}^2 \times B$. 


Different line-shape filters or tools are available in the literature to extract lines and orientation features of a texture image \citep{khan2016optimized}. 2-D Gabor filters have been found more suitable filter bank to extract biological-like textural features of simple cells in the mammalian visual system \citep{Kamarainen2006Feature}. Thus, a Gabor filter is ideal for calculating multi-orientation texture features of a mammogram patch. A Gabor function is defined as follows:

\begin{equation}
\label{eq:gabor}
\begin{array}{lcr}
G(x,y,\theta,\mu,\sigma) &=& \frac{1}{2\pi\sigma^2}\exp\left\{-\frac{x^2+y^2}{2\sigma^2}\right\} 
 \exp\left[2\pi i \left(\mu x \cos \theta+\mu y \sin \theta\right) \right],
\end{array}
\end{equation}
where $i = \sqrt{-1}$, $\mu$ is the frequency of the sinusoidal wave, $\theta$ controls the orientation of the function, and $\sigma$ is the standard deviation of the Gaussian envelop. Based upon this Gabor function, a set of Gabor filters can be created for different scales and orientations. 

Here, texture feature extraction for a given mammogram patch image ($I$) is calculated by the real part of a Gabor filter bank with eight different orientations and a fixed scale. Gabor magnitude, $m(x,y)_{Gabor}$, and Gabor orientation, $\theta(x,y)_{Gabor}$, response of each pixel $(x,y)$ are computed as
\begin{equation}
m_{Gabor}(x,y) = min(I(x,y)\ast G(x,y,\theta_{t},\mu,\sigma)) \ \ \textrm{and} \\\nonumber
\end{equation} 
\begin{equation}
\theta_{Gabor}(x,y) = argmin_{t}(I(x,y)\ast G(x,y,\theta_{t},\mu,\sigma)),\nonumber
\end{equation} 
where $*$ means the convolution operation. The direction $\theta_t$ is calculated as follows:
\begin{equation}
\theta_t= \frac{\pi(t-1)}{8}, \textrm{ } t = 1,2,\ldots,8.\nonumber
\end{equation}

\begin{figure}[h!]
	\centering
		\includegraphics[width=.6\textwidth]{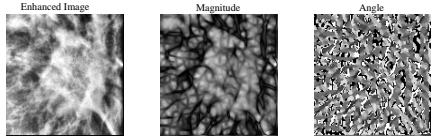}
	\caption{Gabor magnitude and angle image.}
	\label{fig:gabor}
\end{figure}

The features are calculated by varying the values of $\sigma$ and $\mu$.
Fig. \ref{fig:gabor} shows the magnitude and angle image of a mammogram patch. 

We combine HOG with a Gabor filter and name it as Histogram of Oriented Texture (HOT). HOT is computed in the same way as HOG, but $m_{Gabor}(x,y)$ and $\theta_{Gabor}(x,y)$ are used as magnitude and orientation of texture line instead of (\ref{eq:m_hog}) and (\ref{eq:t_hog}), respectively.  

Finally, optimum parameters of the HOT descriptor, for both types of classification, are chosen by experiments. The value of $\sigma$ is varied from one to five to obtain an optimum number.
The value of $\mu$ is computed as $\frac{1}{\sqrt{2\sigma}}$.
In this work, the magnitude image is divided into equal sized $16 \times 16$ cells. Size of a block considered is $2 \times 2$, therefore, $15 \times 15$ overlapped blocks are formed. The orientation range ($0^{0}-180^0$) is quantized into $8$ bins, and therefore, the final length of the resultant HOT descriptor is 7200. 
The length of the HOT descriptor is large, and all features do not have same discrimination capability \citep{liu2002comparative}. Feature selection schemes help to select more appropriate features. This is discussed in Section \ref{sec:featureselection}.

{\color{rev2color}
Next, we compare our proposed HOT descriptor with other works that use a combination of HOG and Gabor filter. In work by \cite{xu2015facial}, authors use a Gabor filter to extract features and HOG to reduce the dimension of the extracted feature vector. However, we use a combination of both to extract features. Comparison with two other works in this area is given in Table \ref{tab:compare_with_HOT}. Apart from the differences discussed until now, we (i.e. in the HOT descriptor) use Discrimination Potentiality for feature selection, which none of the other works use.}

\begin{table}[h]
	\centering
	{\color{rev2color}
	\caption{Comparison of various descriptors using a combination of HOG and Gabor filter}
	\label{tab:compare_with_HOT}
	\smallskip
	
	\begin{tabular}{|l|c|c|c|}
		\hline	
		\multicolumn{1}{|c|}{\textbf{Parameters}} & \textbf{\begin{tabular}[c]{@{}c@{}}HoGG \\ \citep{conde2013hogg}\end{tabular}} & \textbf{\begin{tabular}[c]{@{}c@{}}Gabor-HOG \\ \citep{ouanan2015gabor}\end{tabular}}  & \textbf{HOT} \\ \hline
		\begin{tabular}[c]{@{}l@{}}Number of \\ Orientations\end{tabular} & 9 & 4 & 8 \\ \hline
		\begin{tabular}[c]{@{}l@{}}Number of \\ Bins\end{tabular} & 9 & 9 & 8 \\ \hline
		\begin{tabular}[c]{@{}l@{}}Number of Cells \\ to Divide Image\end{tabular} 
		& 4 $\times$ 8 & \begin{tabular}[c]{@{}c@{}}Rectangular Cells \\ (Number Not Given)\end{tabular}  & 16 $\times$ 16 \\ \hline
		\begin{tabular}[c]{@{}l@{}}Overlapped \\ Image Area\end{tabular} 
		& 3 $\times$ 7 & \begin{tabular}[c]{@{}c@{}}Half of the \\ Image\end{tabular}  & 15 $\times$ 15 \\ \hline
		\begin{tabular}[c]{@{}l@{}}Feature Vector \\ Length\end{tabular} & 756 & 81 & 7200 \\ \hline
		Application Area & 
		\begin{tabular}[c]{@{}c@{}}Human \\ Detection for \\ Surveillance\end{tabular} &
		\begin{tabular}[c]{@{}c@{}}Face \\ Recognition\end{tabular} &  \begin{tabular}[c]{@{}c@{}}Mammogram \\ Patch \\ Classification\end{tabular} \\ \hline
	\end{tabular}
}
\end{table}


\subsubsection{Pass Band - Discrete Cosine Transform (PB-DCT)}\label{sec:DCT}
2D Discrete Cosine Transform (DCT) transforms images into frequency representation from the spatial form.  It also provides energy compaction, that helps to reduce the information redundancy by retaining only a few coefficients. DCT coefficients of $I(x,y)$ image of $M\times N$ size are calculated as follows:

\begin{eqnarray}
F(u,v)& = & \frac{1}{\sqrt{MN}}\alpha(u)\alpha(v)\sum_{x=0}^{M-1}\sum_{y=0}^{N-1}I(x,y)\times \\ \nonumber
&& \cos\left(\frac{(2x+1)u\pi}{2M} \right)\times cos\left(\frac{(2y+1)v\pi}{2N} \right)
 \textrm{ with} \\\nonumber
&&{\textrm{   } u = 0,1,2,\ldots,M,  \textrm{   }  v = 0,1,2,\ldots,N,}\nonumber
\end{eqnarray}
where $\alpha(\omega)$ is defined by

\begin{equation}
\alpha(\omega) = \left\{\begin{array}{cc}
\frac{1}{\sqrt{2}}&\omega = 0\\\nonumber
1 & \textrm{otherwise}.
\end{array}
\right.
\end{equation}

DCT based various feature extraction and compression techniques have been proposed in literature \citep{Dabbaghchian2010Feature}. Usually, DCT features are formed by selecting the most prominent and discriminating coefficients based upon some criterion \citep{Dabbaghchian2010Feature}. The DCT coefficients can be divided into three sets, low frequencies, middle frequencies and high frequencies. 
Low frequencies are correlated with illumination conditions, middle frequencies represent texture features, while high frequencies represent small variance or noise. Illumination and texture properties are important for mammogram patch classification. Therefore, this work uses low and middle coefficients to form the descriptor (the Pass Band - Discrete Cosine  descriptor or abbreviated as PB-DCT). Finally, the more discriminate DCT coefficients are selected based upon a discrimination criterion, which is discussed in the next section. 

 
 \subsection{Feature Selection with Discrimination Potentiality}
\label{sec:featureselection}
All features do not have the same ability to discriminate various classes (normal-abnormal and benign-malignant) \citep{liu2002comparative}, and they do not increase the accuracy based on available information for each class. Therefore, it is necessary to eliminate irrelevant features and select the most discriminative features among a given set of features \citep{kao2010local}. Determining features to improve accuracy as well as reduce searching time is a difficult task. As discussed in Section \ref{sec:lit}, feature subset selection techniques have been found to be more suitable for mammogram patch classification as compared to dimension reduction techniques.

{\color{rev2color}There exist many techniques for feature selection. Some of the common ones are as follows: PCA (Principal Component Analysis) method, Markov blanket method, wrapper methods (e.g., sequential selection algorithm, genetic algorithms etc.), filter methods (e.g., Pearson correlation criteria, mutual information etc.), embedded methods, and statistical measures based methods (e.g., T-test, Kolmogorov-Smirnov test, Kullback-Leibler divergence etc.) \citep{chandrashekar2014survey, zeng2009classification}.

Out of these, wrapper methods, filter methods, and statistical measures based methods are usually used for mammogram patch classification. Wrapper methods are computationally expensive since the number of steps required for obtaining the feature subset are very high. Filter methods sometimes lead to a redundant feature subset, and hence, are not optimal in this sense.  \citep{chandrashekar2014survey}. Thus, we go for statistical measures based methods since they do not have the above discussed drawbacks. These methods also have the advantage in reducing the feature space without significantly degrading the classification performance \citep{dong2009feature}. 

The T-test method is one such method that gives a high score to features that capture the texture and the shape of mammograms \citep{nandi2006genetic}. As earlier, capturing texture is very important to us. Moreover, the T-test method is computationally light, easy to implement, and has been very recently successfully applied in mammogram context \citep{gedik2016new, Jenifer2016contrast}. Thus, we use this feature selection method and show in the results section that this works very well with our proposed descriptors (DP-HOT and DP-PB-DCT). We term it as Discrimination Potentiality (DP) because of its capability in discriminating between the available features. 
}

The discrimination potentiality $DP_{k}$ of the $k^{\textrm{th}}$ feature between two classes (a and b) is computed from a given training set as follows:
\begin{equation}
DP_{k}= \frac{\mu_{a,k}-\mu_{b,k}}{\sqrt{\frac{\delta_{a,k}^2}{n_{a}}-\frac{\delta_{b,k}^2}{n_{b}}}}, \nonumber
\end{equation}
where $\mu_{a,k}$, $\mu_{b,k}$, and $\delta_{a,k}$, $\delta_{b,k}$, are mean and standard deviation values of the $k^\textrm{th}$ feature for $a$ and $b$ classes, respectively. $n_{a}$ and $n_{b}$ are the number of mammogram patches for $a$ and $b$ classes, respectively. A high value of DP means high discrimination ability of the corresponding feature \citep{Jenifer2016contrast}. 

All features (columns) of the feature matrix are arranged in descending order of their $DP$ value. 
Initially, first five features with highest $DP$ values are chosen for classification accuracy. Then, classification accuracy is calculated by adding features, with next higher value of $DP$, one by one until we get the highest accuracy. The optimum subset of features corresponding to the highest accuracy is selected as the final descriptor.

\section{Experimental Results}
\label{sec:exp}
Experiments are carried out in $\textrm{MATLAB}^{\textregistered}$ on a machine with Intel Quad Core processor @2.83GHz and 8GB RAM. As discussed earlier, mammogram patches are taken from the IRMA database (the MIAS and DDSM datasets). All images from this database (density wise), as given earlier in Table 1, are used. The size of each mammogram patch is $128 \times 128$.  

%
%
%
%
{\color{rev1color}
We {\it first} use two-fold cross validation, where the dataset is randomly divided into two equal parts. One part is used for training and the other is used for testing. Then, the two parts are swapped. That is, the one used for training earlier is now used for testing, and the one used for testing earlier is now used for training. At the end of this exercise, average performance is saved. {\it Finally}, we repeat two-fold cross validation ten times so as to remove any bias related to the division of the dataset. 
}

The performance of our system (and comparative systems) is evaluated by standard metrics of sensitivity, specificity, accuracy {\color{rev1color}and AUC (Area Under the ROC Curve).} 

{Sensitivity} is computed as the number of true positive cases over the number of actual positive cases. It is represented as follows:
\begin{equation}
Sensitivity = \frac{TP}{TP+FN}(\%),\nonumber
\end{equation}
where TP means True Positive cases and FN means False Negative cases.

{Specificity} is computed as the number of true negative cases over the number of actual negative cases. It is represented as follows:
\begin{equation}
Specificity = \frac{TN}{FP+TN}(\%),\nonumber
\end{equation}
where TN means True Negative cases and FP means False Positive cases.

{Accuracy} is computed as the number of correct classifications over the number of given cases. It is represented as follows:
\begin{equation}
Accuracy = \frac{TP+TN}{TP+FN+FP+TN}(\%).\nonumber
\end{equation}

{\color{rev1color}
AUC (Area Under the ROC Curve) provides a measure of the overall performance of the classifier, i.e. larger the area, better the classification. It is calculated from the trapezoidal rule as below \citep{yeh2002using}.
\begin{equation}
AUC =\frac{1}{2}\sum_{k=1}^{n}((Spec(k)-Spec(k+1))*(Sens(k)+Sens(k+1)))(\%), \nonumber
\end{equation}
where $n$ is the total number of test cases, and $Spec(k)$ and $Sens(k)$ are Specificity and Sensitivity for the $k^{th}$ test case, respectively. 
In the case of an ideal classification system, the value of all the four metrics should be close to 100\%.
}

This section has three subparts. In Section \ref{dphotperform}, experiments related to finding the optimum parameters of the Histogram of Oriented Texture (HOT) descriptor with Discrimination Potentiality (DP), from now on referred as DP-HOT, are given. Similar experimental results for the Pass Band - Discrete Cosine Transform (PB-DCT) descriptor with Discrimination Potentiality (DP), from now on referred as DP-PB-DCT, are given in Section \ref{dpdctperform}. Finally, in Section \ref{bothperform}, the performance of both our proposed descriptors, DP-HOT and DP-PB-DCT, is compared with the performance of the existing (and popular) descriptors. 

{\color{rev1color} 
For fixing the parameters, only the training data is used ($i.e.$ 50\%, as we use two-fold cross validation) and the validation data for testing is kept blind. 
}

\subsection{Performance of DP-HOT}\label{dphotperform}
Firstly, experiments are performed to find optimum parameters of Gabor filter for all the classes. Performance parameters are calculated by varying the value of $\sigma$ from one to five to obtain suitable scale for each density class. The best accuracy obtained by varying $\sigma$ is mentioned here. 

\begin{figure*}[h]
	\centering
	\begin{tabular}{c}
		\includegraphics[width = 1\textwidth]{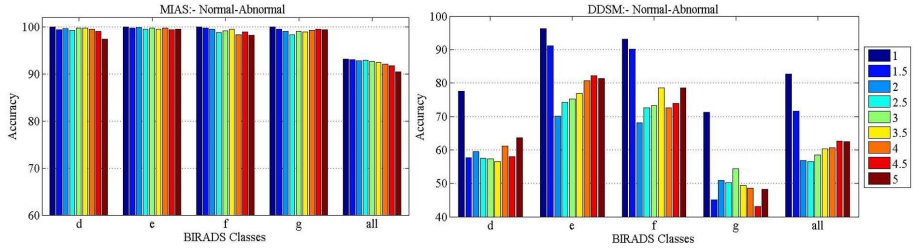}

	\end{tabular}
	\caption{\label{fig:sigmahotnormal}Comparison of normal-abnormal classification accuracies obtained by varying the value of $\sigma$ from one to five for each individual BIRADS class.}
\end{figure*} 

Fig. \ref{fig:sigmahotnormal} compares normal-abnormal classification accuracy for different values of $\sigma$ for each class. It is observed that the DP-HOT descriptor achieves approximately 100\% accuracy for all values of $\sigma$ for all the classes of the MIAS dataset. The maximum normal-abnormal classification accuracy of all the classes combined is achieved with $\sigma$ as one. It is difficult to infer the optimum value of $\sigma$ by observing the bar chart of the MIAS dataset only. In case of the DDSM dataset, the DP-HOT descriptor with $\sigma$ as one gives the best accuracy for all the classes individually as well as combined. For classes $e$ and $f$, DP-HOT achieves an accuracy of around 95\%, while it does not achieve good accuracy for classes $d$ and $g$ (around 70\%). 

\begin{figure*}[h]
	\centering
	\begin{tabular}{c}
				\includegraphics[width = 1\textwidth]{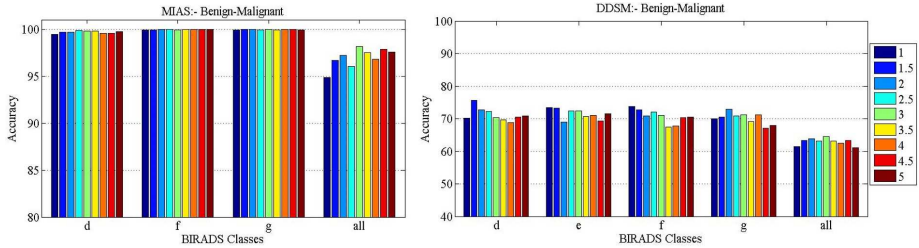}

	\end{tabular}
	\caption{\label{fig:sigmahot}Comparison of benign-malignant classification accuracies obtained by varying the value of $\sigma$ from one to five for each individual BIRADS class.}
\end{figure*}
Fig. \ref{fig:sigmahot} compares benign-malignant classification accuracy for different values of $\sigma$ for each class. The DP-HOT descriptor again achieves approximately 100\% accuracy for all values of $\sigma$ for all the classes of the MIAS dataset. The maximum benign-malignant classification accuracy for all the classes combined is achieved with  $\sigma$ as 3. For the DDSM dataset, the DP-HOT descriptor performs equally for all $\sigma$ for all the classes (around 70\%).


First observation is that DP-HOT does not perform very well for both types of classification (normal-abnormal and benign-malignant) for the DDSM dataset (although it does perform well for MIAS). The reason, as discussed in Section \ref{sec:lit}, is that texture information plays a big role in both types of classifications (normal-abnormal and benign-malignant), and DP-HOT does not capture that well for the DDSM dataset. The DP-PB-DCT descriptor overcomes this drawback to a great extent. Another observation is that classification accuracy for an individual class is better than combined. This is intuitive since images in a class are similar.


Table \ref{tab:hot} lists the feature length used for both types of classification (normal-abnormal and benign-malignant) done on both the datasets (MIAS and DDSM) for each density class. 

\begin{table*}[h]
\caption{\label{tab:hot} 
Feature length for the DP-HOT descriptor.} \smallskip
\begin{tabular*}{\hsize}{@{\extracolsep{\fill}}|r|r|r|@{}}
\hline
\multirow{1}{*}{BIRADS Class}& \multicolumn{1}{c|}{MIAS: Feature Length}	&	\multicolumn{1}{c|}{DDSM: Feature Length} \\\hline	
\multicolumn{3}{|c|}{\textbf{Normal-Abnormal}}\\\hline	
$d$&2655&210\\
$e$&10&100\\
$f$&30&75 \\
$g$&20&200\\
All&40&140\\\hline	
\multicolumn{3}{|c|}{\textbf{Benign-Malignant}}\\\hline	
$d$&2000&225\\
$e$&All Benign&130\\
$f$&2000&405\\
$g$&1790&170 \\
All& 2000&990\\\hline
 \end{tabular*}
\end{table*}

\subsection{Performance of DP-PB-DCT}\label{dpdctperform}

As in the case of the DP-HOT descriptor, in Table \ref{tab:DP_DCT} we list the feature length used for both types of classification (normal-abnormal and benign-malignant) done on both the datasets (MIAS and DDSM) for each density class when using the DP-PB-DCT descriptor. 
	\begin{table*}[h]
\caption{\label{tab:DP_DCT} 
Feature length for the DP-PB-DCT descriptor.} \smallskip
\begin{tabular*}{\hsize}{@{\extracolsep{\fill}}|r|r|r|@{}}									\hline
\multirow{1}{*}{BIRADS Class}& \multicolumn{1}{c|}{MIAS: Feature Length}	&	\multicolumn{1}{c|}{DDSM: Feature Length} \\\hline	
\multicolumn{3}{|c|}{\textbf{Normal-Abnormal}}\\\hline	
$d$&5&4000\\
$e$&5&4000\\
$f$&5&995\\
$g$&5&1315\\
All&5&430\\\hline
\multicolumn{3}{|c|}{\textbf{Benign-Malignant}}\\\hline	
$d$&5&4350\\ 
$e$&5&4005 \\ 
$f$&5&3995 \\ 
$g$&5&4625\\ 
All&5&5065 \\\hline
 \end{tabular*}
\end{table*}

\begin{figure*}[h]
	\centering
	\begin{tabular}{cc}
				\includegraphics[width= .45\textwidth]{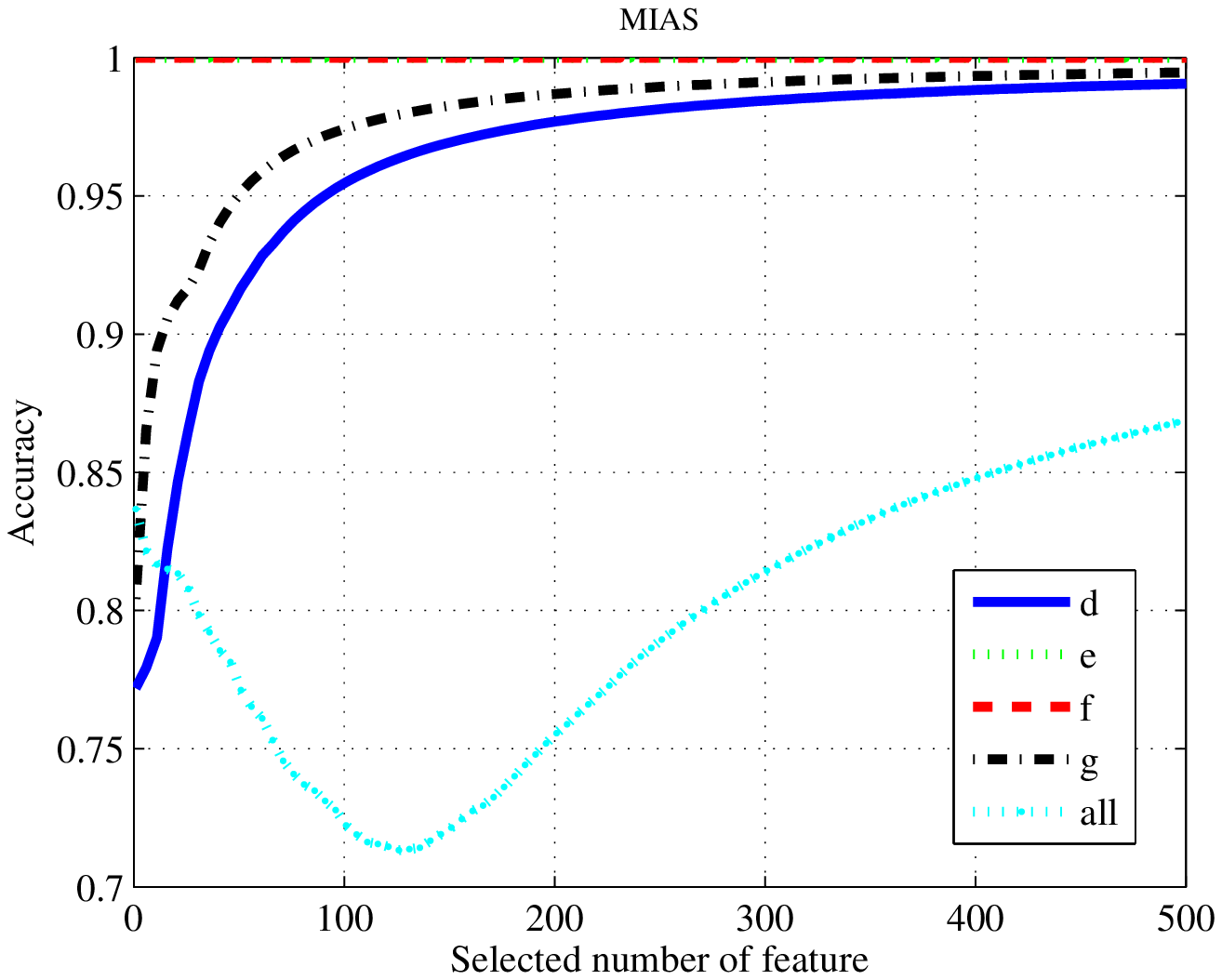}&	
				\includegraphics[width = .45\textwidth]{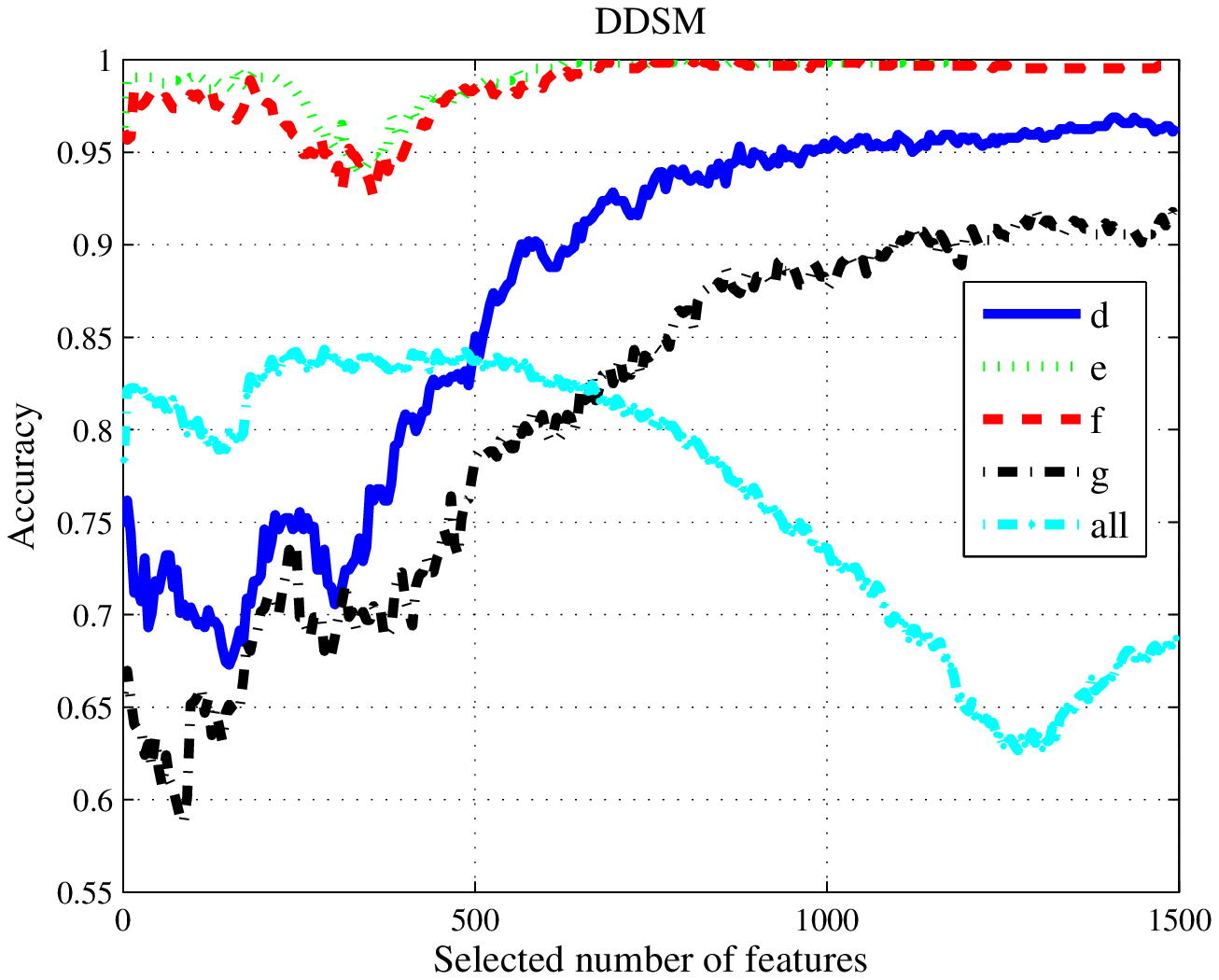}\\
					(MIAS)&(DDSM)\\
				\multicolumn{2}{c}{Normal-Abnormal}\\
			\includegraphics[width= .45\textwidth]{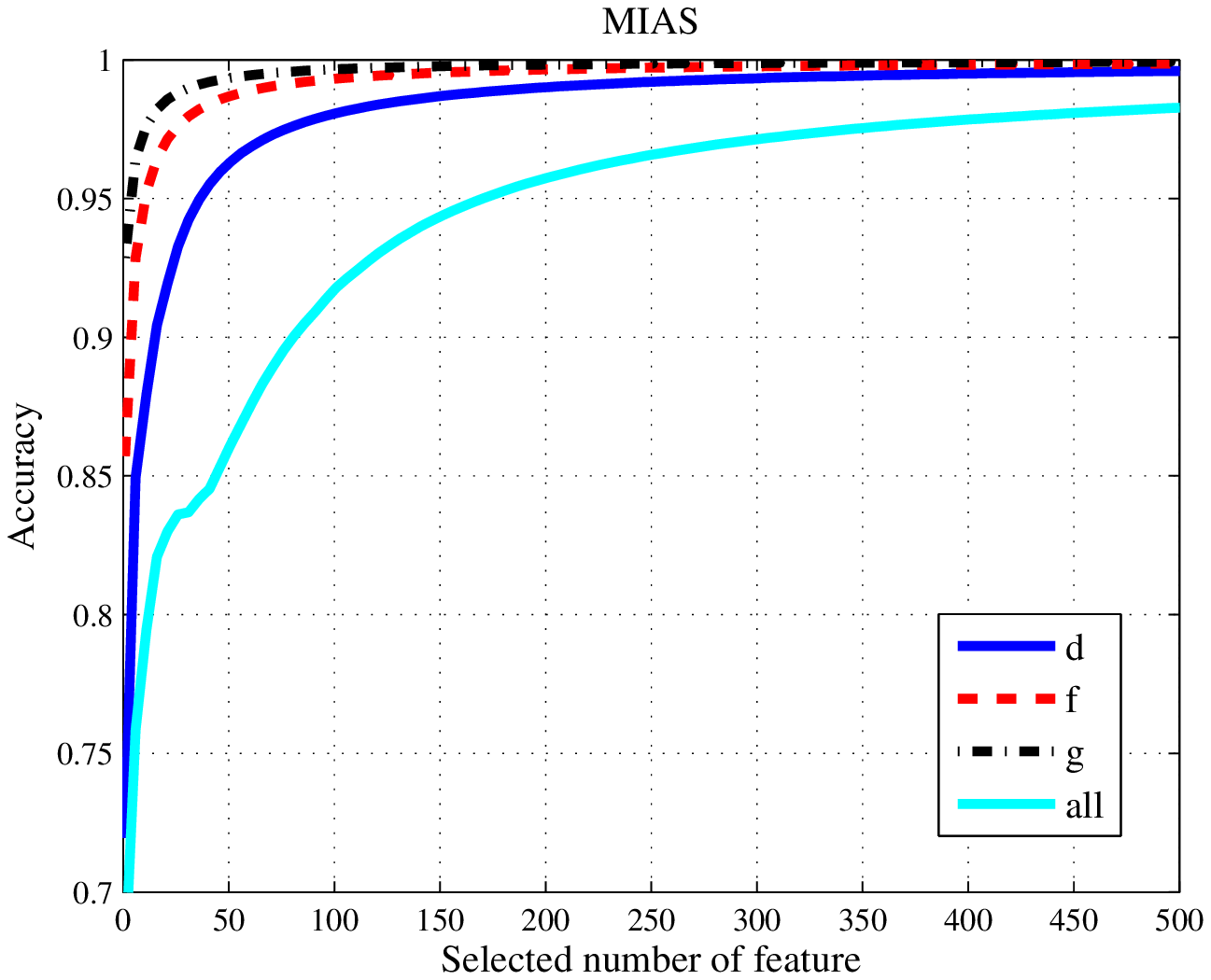}&
			\includegraphics[width = .45\textwidth]{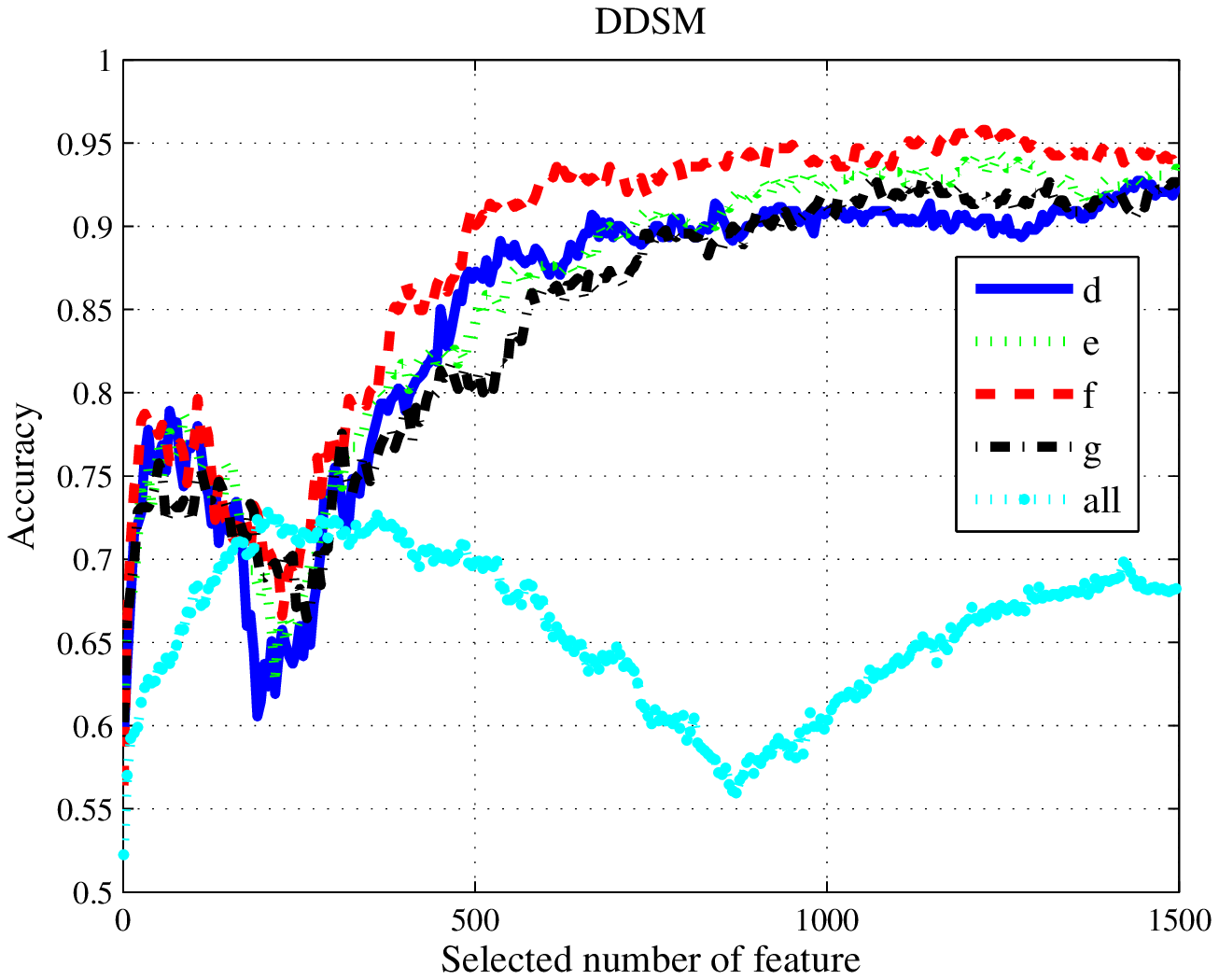}\\
		(MIAS)&(DDSM)\\
		\multicolumn{2}{c}{Benign-Malignant}
	\end{tabular}
	\caption{\label{fig:benignperfomance}Performance accuracy against the number of DP-PB-DCT features.}
\end{figure*}

Since, the selection of features in the DP-PB-DCT descriptor is more critical, we do further feature selection analysis here. Fig. \ref{fig:benignperfomance} compares accuracy against the number of features for normal-abnormal and benign-malignant classification for each density class separately and combined. As in the case of DP-HOT, the performance for individual density is better than combined. Moreover, multiple points with the high classification accuracy are observed.

\subsection{Comparison with Other Techniques}\label{bothperform}
The performance of the two proposed descriptors is compared with some related descriptors such as Zernike moment \citep{tahmasbi2011classification}, {\color{rev1color}MLPQ \citep{nanni2012very}, GRsca \citep{nanni2013different},} WGLCM \citep{beura2015mammogram}, LCP \citep{ergin2014new} and HOG \citep{ergin2014new} for each density class separately as well as combined. {\color{rev1color}The performance of each descriptor is evaluated in the same experimental setup including the use of the same testing protocol.} 

The parameters of each of these descriptors are selected as given in the literature to achieve the best performance for mammogram patch classification. These parameters are summarized below. 

The Zernike moment descriptor is computed by dividing a mammogram patch into $4 \times 4$ blocks. $120$ Zernike moments are used for obtaining the final mammogram patch descriptor. 
Therefore, the length of the feature vector is $4 \times 4 \times 120$ \citep{tahmasbi2011classification}.  

{\color{rev1color}
The MLPQ descriptor is computed by concatenating different Local Phase Quantization (LPQ) descriptors. Each LPQ descriptor is obtained by varying values of three different parameters. This includes filter size ($r$ with values $1, 3, 5$), the scalar frequency ($a$ with values $0.8, 1.0, 1.2, 1.4, 1.6$), and the correlation coefficient ($\rho$ with values $0.75, 0.95, 1.15, 1.35, 1.55, 1.75, 1.95$). Each LPQ descriptor's length is standard ($256$). Therefore, the length of the feature vector is $3 \times 5 \times 7 \times 256$ \citep{nanni2012very}.  

The GRsca descriptor extracts features from the three gray-level run length co-occurrence matrices corresponding to the original image and two filtered images (using a filter of size $3$ and $5$). These images are divided into four blocks. Features are obtained from co-occurrence matrix and these four blocks. A set of ten different descriptors is calculated at four different orientations and two different distances. Therefore, the length of the feature vector is $3 \times 5 \times 10 \times 4 \times 2$ \citep{nanni2013different}.	


}

The Wavelet Gray Level Co-occurrence Matrix (WGLCM) descriptor is computed by decomposing the image into one approximation. Detail coefficients up to two levels with a wavelet filter are used. For each level, three decompositions (along horizontal, vertical and diagonal directions) are obtained. 
The normalized GLCM (NGLCM) matrices of all 
detail coefficients are calculated 
in four directions ($0^\circ$, $45^\circ$, $90^\circ$, and $135^\circ$) with a single displacement. Contrast, homogeneity, energy, and correlation statistical properties of all NGLCM matrices are calculated and concatenated into a vector. Therefore, the length of the feature vector is $2 \times 3 \times 4 \times 4$ \citep{beura2015mammogram}.

Local Configure Pattern (LCP) is a modification of Local Binary Pattern (LBP). Here, first, the weights associated with intensities of neighboring pixels are used to linearly reconstruct the central pixel intensity. Then, the error between the central pixel and its neighbor is minimized. In this work, LCP images are computed for radius $1$ to $5$. 
Each LCP image is divided into $4 \times 4$ blocks. The histogram of each block is concatenated with $58$ bins, and the result is used as a mammogram patch descriptor. Therefore, the length of the feature vector is $5 \times 4 \times4 \times 58$ \citep{ergin2014new}. 

Histogram of Gradients (HOG) is calculated using $16 \times 16$ cell partitions. The size of the block considered is $2 \times 2$, thus, $15 \times 15$ overlapped blocks are formed. The orientation range is quantized into $8$ bins. Therefore, the length of the feature vector is $2 \times 2 \times 15 \times 15 \times 8$ \citep{ergin2014new}. 

The optimum parameters of DP-HOT  and DP-PB-DCT are selected based upon experiments as discussed in the previous subsections. The length of each descriptor is different and the appropriate feature set for each descriptor is selected based upon the DP values as described earlier. 


\subsubsection{Classification Results}\label{subSubSecResults}
Table \ref{tab:normal_DDSM1} compares sensitivity, specificity, accuracy {\color{rev1color}and AUC} for normal-abnormal classification with all the above descriptors on the MIAS and DDSM datasets. The performance parameters for all the descriptors are provided for each density class separately as well as combined. {\color{rev1color}The best performing systems are highlighted in bold.} 

\begin{table}[!htbp]
	\caption{\label{tab:normal_DDSM1} 
		Mammogram patch classification results as normal-abnormal for the MIAS and DDSM datasets.} \smallskip
	\begin{tabu}{|p{2.5cm}|r|r|r|r|>{\color{rev1color}}r|r|r|r|>{\color{rev1color}}r|}                                                
		\cline{1-10} 
		\multirow{4}{*}{Descriptor}& \multirow{2}{*}{\rotatebox[origin=c]{90}{BIRADS~}}&    \multicolumn{4}{c|}{\multirow{1}{*}{MIAS}}    &    \multicolumn{4}{c|}{\multirow{1}{*}{DDSM}}        \\ 
		
		\cline{3-10}      
		& & & & & & & & & \\                                           
		& &Sens.     &    Spec.      &    Acc. & AUC &    Sens.     &    Spec. &    Acc. & AUC \\ 
		   
		& & & & & & & & & \\ \cline{1-10}
		\multirow{5}{1.4cm}{Zernike \citep{tahmasbi2011classification}}&    
		$d$&99.12&93.59&97.32 &90.28 & 90.01&    68.64&    83.32 & 91.69\\
		&$e$&94.98&    99.63&98.81 &\textbf{100} &99.78&    85.12&    95.86 & \textbf{100}\\
		&$f$&92.86&99.88&97.29& \textbf{100}  &    99.56&    95.91&    98.46 & \textbf{100}\\
		&$g$&91.31 &97.87&95.42& 75.82 & 83.32&    36.72&    68.63 & 81.89\\
		&All&75.92 &91.70    &85.39 &85.11 &  \textbf{91.90}&61.99&    82.95 & 82.03\\ \cline{1-10}
		
		\rowfont{\color{rev1color}}
		\multirow{5}{1.4cm}{MLPQ \citep{nanni2012very}}	
		&$d$&89.21&	75.00&	84.70 &90.28	& 80.29&73.51&	78.16 & 89.71\\ 
		\rowfont{\color{rev1color}}
		&$e$& \textbf{100}			&96.30		&96.96&	\textbf{100}& 	99.97&99.97&	99.97 &\textbf{100} \\ 
		\rowfont{\color{rev1color}}
		&$f$&\textbf{100}&	\textbf{100}&\textbf{100} &\textbf{100}	& 99.91&	99.91&	99.91 & \textbf{100}\\ 
		\rowfont{\color{rev1color}}
		&$g$&68.75&	92.31&	83.10& 90.66&83.14&	54.06&	73.97 & 81.80\\ 
		\rowfont{\color{rev1color}}
		&All&	72.60&88.50&	82.14& 88.15	& 	86.07&	88.14&	86.69 & 93.59\\ \cline{1-10}
		
		\rowfont{\color{rev1color}}
		\multirow{5}{1.4cm}{GRsca \citep{nanni2013different}}	
		&$d$&76.92&	70.00&	74.59 &85.90	& 86.03&79.53&	84.00 & 88.23\\ 
		\rowfont{\color{rev1color}}
		&$e$&\textbf{100}		&\textbf{100}		&\textbf{100}&\textbf{100}& 	\textbf{100}&	\textbf{100}&\textbf{100} &\textbf{100} \\ 
		\rowfont{\color{rev1color}}
		&$f$&\textbf{100}&	\textbf{100}&\textbf{100} &\textbf{100}	& \textbf{100}&	\textbf{100}&\textbf{100} & \textbf{100}\\ 
		\rowfont{\color{rev1color}}
		&$g$&66.96&	\textbf{100}&87.86& 84.13&83.41&	63.85&	77.24 & 82.26\\ 
		\rowfont{\color{rev1color}}
		&All&68.95&98.79&86.85& 91.26	& 	88.44&	88.64&88.50 & \textbf{95.48}\\ \cline{1-10}

		\multirow{5}{1.4cm}{WGLCM \citep{beura2015mammogram}}&
		$d$&92.15&    50.83&    78.71&73.61 & 85.54&    84.92&    85.34 & 94.38\\
		&$e$&90.91&    \textbf{100}&    98.40& \textbf{100}& 99.84&    99.98&    99.87 & \textbf{100}\\
		&$f$&\textbf{100}&    \textbf{100}&    \textbf{100} &\textbf{100} & 99.70&    \textbf{100}&    99.79 & \textbf{100}\\
		&$g$&60.96&    \textbf{100}&    85.93& 80.77 & 84.94&    62.25&    77.78 & 85.81\\
		&All&64.72&    \textbf{100}&    85.89&93.41 & 86.24&    93.21&    88.33 & \textbf{94.97}\\ \cline{1-10}
		
		\multirow{5}{1.4cm}{LCP \citep{ergin2014new}}&                    
		$d$&92.31&    66.67&    83.63& 80.56& 87.99&    89.07&    88.32    & 90.39   \\
		&$e$&\textbf{100}&    \textbf{100}&    \textbf{100}&\textbf{100} & \textbf{100}&    \textbf{100}&    \textbf{100}    & \textbf{100}   \\
		&$f$&\textbf{100}&    \textbf{100}&    \textbf{100}&\textbf{100} & \textbf{100}&    \textbf{100}&    \textbf{100} & \textbf{100}\\
		&$g$&73.21&    96.15&    87.74&64.42 & 86&    69.10&    80.67 & 74.93\\
		&All&75&    97.78    &88.67&85.70 & 87.80&    \textbf{93.39}&    \textbf{89.47} & 92.88\\ \cline{1-10}
		
		\multirow{5}{1.4cm}{HOG \citep{ergin2014new}}                                                        
		&$d$&    88.46    &    91.67    &    89.47& \textbf{100} & 84.36&    78.62&    82.56 & 87.58 \\
		&$e$&    \textbf{100}    &    \textbf{100}    &    \textbf{100}& \textbf{100}& 99.57&    94.64&    98.25 &98.79 \\
		&$f$&    \textbf{100}    &    95.83    &    97.36& \textbf{100}& 98.23&    88.73&    95.36 &97.67 \\
		&$g$&    85.71    &    \textbf{100}    &    95&\textbf{100} & 82.22&    56.52&    74.12 &77.70 \\
		&All&    70    &    87.78    &    80.67&85.85 & 87.69&    77.95&    84.77 & 88.32\\ \cline{1-10}    
		\multirow{5}{1.4cm}{\textbf{DP-HOT}}
		&$d$&    \textbf{100}    &    \textbf{100}    &    \textbf{100}&\textbf{100} & 83.44&    64.69&    77.57 &91.20 \\
		&$e$&    \textbf{100}    &    \textbf{100}    &    \textbf{100}&\textbf{100} & 96.96&    95.24&    96.50 &\textbf{100} \\
		&$f$&    \textbf{100}&    \textbf{100}    &    \textbf{100}& \textbf{100}& 94.69&    89.75&    93.20 & \textbf{100} \\
		&$g$&    \textbf{100}    &    \textbf{100}    &    \textbf{100}&\textbf{100} &77.78&    57.06&    71.24 &76.05 \\
		&All&    87    &    98    &    93& 97.26& 84.80&    77.30&    82.56 & 86.21\\ \cline{1-10}    
		
		\multirow{5}{1.4cm}{\textbf{DP-PB-DCT}}                                            &$d$&\textbf{100}&    \textbf{100}&    \textbf{100}& \textbf{100} & {\textbf{98.19}}&{\textbf{94.01}}&    {\textbf{96.88}}   &\textbf{98.39} \\
		&$e$&    \textbf{100}&    \textbf{100}&    \textbf{100} & \textbf{100}&     {\textbf{100}}&{\textbf{100}}&    {\textbf{100}} &\textbf{100}\\
		&$f$&    \textbf{100}&    \textbf{100}&    \textbf{100}&\textbf{100} &     {\textbf{100}}&    {\textbf{100}}&    {\textbf{100}} &\textbf{100}\\
		&$g$&    \textbf{100}&    \textbf{100}&    \textbf{100}&  \textbf{100}   &{\textbf{97.78}}&    \textbf{80.70}&    \textbf{92.39} & \textbf{98.68}\\
		&All&\textbf{97.18}&98.89&\textbf{97.33}&  \textbf{100}   & 87.18&    79.37&    {84.84} & 92.20\\ \cline{1-10}
	
	\end{tabu}
\end{table}

\begin{table}[!htbp]
	\caption{\label{tab:benign} 
		Mammogram patch classification results as benign-malignant for the MIAS and DDSM datasets.} \smallskip
	\begin{tabu}{|p{2.5cm}|r|r|r|r|>{\color{rev1color}}r|r|r|r|>{\color{rev1color}}r|}												
		\cline{1-10}
		\multirow{4}{*}{Descriptor}&
		\multirow{2}{*}{\rotatebox[origin=c]{90}{BIRADS~}}&	\multicolumn{4}{c|}{\multirow{1}{*}{MIAS}}	&	
		\multicolumn{4}{c|}{\multirow{1}{*}{DDSM}}	\\	\cline{3-10}
		
		& & & & & & & & & \\												
		& &Sens.     &    Spec.      &    Acc. & AUC &    Sens.     &    Spec. &    Acc. & AUC \\
		& & & & & & & & & \\ \cline{1-10}

		\multirow{5}{1.4cm}{Zernike \citep{tahmasbi2011classification}}&	
		
		$d$&		93.23&	97.05&	95.39& 54.76& 60.36&	54.34&	57.37 &56.78 \\
		&$e$&	-	&  -        &- &- &60.09&	60.34&	60.22 & 58.39\\
		&$f$&97.26&	99.83&	98.73&\textbf{100} & 	58.62&	56.02&	57.30 &63.26 \\
		&$g$&95.61&	98.96&	97.62& \textbf{100}& 57.96&	70.98&	64.44 & 57.22\\
		&All&80.47&	84.15&	82.44&72.77 & 56.92&	55.11&	56.02 & 53.87\\ \cline{1-10}
		
		\rowfont{\color{rev1color}}
		\multirow{5}{1.4cm}{MLPQ \citep{nanni2012very}}	
		&$d$&71.43&	75.00&	71.79 &96.43	& 59.59&	57.43&	58.50 & 62.70\\ 
		\rowfont{\color{rev1color}}
		&$e$&	-			&-			&-&	-& 61.72&	59.36&	60.55 &63.24\\
		\rowfont{\color{rev1color}}
		&$f$&96.43&	97.62&96.94 &\textbf{100}	& 49.79&64.76&	57.31 & 59.18\\ 
		\rowfont{\color{rev1color}}
		&$g$&100&	83.33&	93.75& 93.34&	\textbf{95.98}&	17.70&	56.67 & 57.66\\ 
		\rowfont{\color{rev1color}}
		&All&	73.96&76.19&	75.00& 86.16	& 	57.40&	54.62&	56.01 & 56.04\\ \cline{1-10}
		
		\rowfont{\color{rev1color}}
		\multirow{5}{1.4cm}{GRsca \citep{nanni2013different}}
		&$d$& 64.29&58.33	&60.26	 &62.86	&66.83 &54.21	&	60.47 &62.01 \\ 
		\rowfont{\color{rev1color}}
		&$e$&	-	&-			&-&	-& 71.09	&53.94	&52.59	 & 66.24\\ 
		\rowfont{\color{rev1color}}
		&$f$& 85.71 &28.57	&61.22 &58.34	& 63.59&56.54	&60.05	 &64.32 \\ 
		\rowfont{\color{rev1color}}
		&$g$& 77.59 &58.28	&69.49  &50 &81.70	&40.27	&60.89	 & 64.58\\ 
		\rowfont{\color{rev1color}}
		&All&96.88	& 35.71 &68.33	& 68.75	& 67.69	&52.63	&	60.15 & 64.21\\ \cline{1-10}
		
		\multirow{5}{1.4cm}{WGLCM \citep{beura2015mammogram}}&
		$d$&35.83&92.86&	67.95 &80.95 & 53.15&	62.99&	58.04 & 60.35\\
		&$e$&-		& -&-			&- & 50.88&	63.79&	57.39 & 56.32\\
		&$f$&45.83&	90.63&	71.43&91.67 & 59.92&	52.01&	55.97 & 59.24\\
		&$g$&47.22&	86.25&	70.83&93.34 & 57.08&	60.71&	58.89 & 57.59\\
		&All&53.81&	74.72&	64.96&70.09 & 53.20&	56.98&	55.09 & 56.59\\ \cline{1-10}
		
		\multirow{5}{1.4cm}{LCP \citep{ergin2014new}}					
		&$d$&99.72 &	99.84 &99.77&\textbf{100} & 51.80&	66.66&	59.18	&	63.38 \\
		&$e$& -		& -&		-	&- &57.89&	62.93&	60.43	&	60.86 \\
		&$f$&\textbf{100} &99.9&99.98	&\textbf{100} & 51.96&	62.24&	57.08 & 60.33\\
		&$g$&99.99	&99.96	 &99.98	&\textbf{100} &56.64&	61.16&	58.89 & 60.04\\
		&All&98.94&95.57&97.37&72.32 & 55.26&	59.33&	57.29 & 57.38\\ \cline{1-10}
		
		\multirow{5}{1.4cm}{HOG \citep{ergin2014new}} 													
		&$d$&		\textbf{100}	&	\textbf{100}	&	\textbf{100} &95.24	&	64.86&	65.30&	65.08 & 66.27\\ 
		&$e$&- &-		&-		&- &		59.65&	65.52&	62.61 & 70.01\\ 
		&$f$&	\textbf{100}	&	\textbf{100}	&	\textbf{100}& \textbf{100} &	66.95&	65.32&	66.14 & 67.92\\ 
		&$g$&	\textbf{100}	&	\textbf{100}	&	\textbf{100}& \textbf{100} &		63.27&	66.07&	64.67 & 68.07\\ 
		&All&80.67	&	92.86	&	87.5 & 94.20	&54.93&	61.89&	58.40 & 65.06\\ \cline{1-10}
		
		\multirow{5}{1.4cm}{\textbf{DP-HOT}}
		&$d$&	\textbf{100}	&	\textbf{100}	&	\textbf{100} & \textbf{100}&72.07&	68.49&	70.29 & 80.44\\
		&$e$&	 -	&	 -	&	- &- & 70.18&	75&	72.61 & 74.92\\
		&$f$&	\textbf{100}	&	\textbf{100}	&	\textbf{100}& \textbf{100} &72.66&	69.31&	71.02  & 83.91\\
		&$g$&	\textbf{100}	&	\textbf{100}	&	\textbf{100}&\textbf{100} &73.45&	68.75&	71.11  & 81.09\\
		&All&	98.83	&	97.57	&	98.24 & \textbf{100}&64.56&	64.67&	64.61 & 68.89\\ \cline{1-10}
		
		\multirow{5}{1.4cm}{\textbf{DP-PB-DCT}}		
		&$d$&\textbf{100}&	\textbf{100}&	\textbf{100} &\textbf{100}	& \textbf{96.85}&	\textbf{94.53}&	\textbf{95.69} & \textbf{98.13}\\ 
		&$e$&	-			&-			&-&	-& 	\textbf{96.05}&	\textbf{93.53}&	\textbf{94.78} &\textbf{98.78} \\ 
		&$f$&\textbf{100}&	\textbf{100}&\textbf{100} &\textbf{100}	& \textbf{97.35}&	\textbf{96.45}&	\textbf{96.90} & \textbf{98.87}\\ 
		&$g$&\textbf{100}&	\textbf{100}&	\textbf{100}& \textbf{100}&	91.59&	\textbf{95.98}&	\textbf{93.78} & \textbf{99}\\ 
		&All&	\textbf{100}&\textbf{100}&	\textbf{100}& \textbf{100}	& 	\textbf{73.53}&	\textbf{71.33}&	\textbf{72.43} & \textbf{82.93}\\ \cline{1-10}
		
	\end{tabu}
\end{table}	


For the MIAS dataset, both the proposed descriptors (DP-HOT and DP-PB-HOT) achieve near 100\% sensitivity, specificity, accuracy, {\color{rev1color}and AUC} for all the classes. This is better than the {\color{rev1color}six standard descriptors (Zernike moment, MLPQ, GRsca, WGLCM, LCP, and HOG).} 

For the DDSM dataset, although our DP-HOT descriptor performs almost as badly as {\color{rev1color}the other six descriptors} for all the classes (as low as around 65\% for one performance parameter), DP-PB-DCT performs extremely well for all the classes (more than 92\% for most performance parameters), which is better than the {\color{rev1color}six standard descriptors. All the eight descriptors (the six standard and the two new)} perform well for classes $e$ and $f$ but have a dip in the performance for classes $d$ and $g$. 
This is because for the DDSM dataset, texture discrimination for $e$ and $f$ classes is better than for $d$ and $g$ classes, which is crucial in normal-abnormal classification. 

Table \ref{tab:benign} compares performance parameters for benign-malignant classification with all the above {\color{rev1color}eight descriptors} for all the classes (and combined) of the MIAS and DDSM datasets. {\color{rev1color}As earlier, the best performing systems are highlighted in bold.} The results are similar to those for normal-abnormal classification. Note that in the MIAS dataset, the $e$ class has all benign images, so, classification was not performed for this. The corresponding rows in this table are left empty.

For the MIAS dataset, both the proposed descriptors (DP-HOT and DP-PB-DCT) perform slightly better than the {\color{rev1color}six standard descriptors} for all the classes (achieve near 100\% sensitivity, specificity, accuracy, {\color{rev1color}and AUC}).

For the DDSM dataset, DP-HOT is slightly better than the existing descriptors (around 70\% for all performance parameters), while DP-PB-DCT is much better than the {\color{rev1color}six standard descriptors} for all the classes (more than 92\% for most performance parameters).

To summarize, as mentioned in Section \ref{sec:lit} as well as Section \ref{dphotperform}, capturing texture information is of utmost importance for mammogram patch classification (both normal-abnormal and benign-malignant). In general, DP-HOT captures this texture information slightly better than the {\color{rev1color}six standard descriptors}, and hence, it performs slightly better than these. DP-PB-DCT captures texture information best, and hence, performs much better than all the others.\\



\section{Conclusion and Future Work}
\label{sec:summary}
{\color{rev2color}We propose a variant of HOG and Gabor filter combination, which we call Histogram of Oriented Texture (HOT) for mammogram patch classification.} We also revisit the Pass Band - Discrete Cosine Transform (PB-DCT) descriptor for the same. The feature selection technique of Discrimination Potentiality (DP) is used with the above two descriptors for mammogram patch classification. This results in two new descriptors (DP-HOT and DP-PB-DCT). We consider the density of mammogram patches as a factor for classification (this was not done earlier), and show that this plays an important role in classification.

Our two-stage patch classification system (normal-abnormal and benign-malignant), using the two new descriptors for each density class, is tested on all images of the MIAS and DDSM datasets from the IRMA repository (in literature, experiments have been done only on a subset of these images). We achieve a high classification performance (in terms of specificity, sensitivity, accuracy {\color{rev1color}and AUC}) in an absolute sense as well as in relative sense (compared with the {\color{rev1color}six standard descriptors}). This is due to the fact that textural information in a mammogram patch is crucial for classification, and our descriptors capture that well. 

Future work here involves developing a Computer-Aided Diagnosis (CAD) application for full mammogram using our framework. This will aid radiologists for more accurate diagnosis of breast cancer using only the information about mammogram density. 




\section{Acknowledgment}
The authors would like to thank Prof. Dr.rer.nat. Dipl.-Ing. Thomas Martin Deserno, Department of Medical Information, Division of Image and Data Management, RWTH Aachen, Germany, for providing us the images. 

Thanks to the anonymous reviewers that helped to greatly improve the quality of this manuscript. We would also like to thank the editor handling our manuscript, Prof. Binshan Lin (at Louisiana State University, Shreveport), in giving us the flexibility during revision submissions.
\section*{References}
\bibliographystyle{model5-names} \biboptions{authoryear}
\bibliography{ref}

\end{document}